\documentclass[lettersize,journal]{IEEEtran}
\usepackage{amsmath,amsfonts}
\usepackage{algorithm}
\usepackage{algpseudocode}
\usepackage{array}
\usepackage[caption=false,font=normalsize,labelfont=sf,textfont=sf]{subfig}
\usepackage{textcomp}
\usepackage{stfloats}
\usepackage[hyphens]{url}
\usepackage{verbatim}
\usepackage{graphicx}
\usepackage{cite}
\usepackage{setspace}
\usepackage{xcolor}
\usepackage{url}
\usepackage[T1]{fontenc}
\usepackage{tabularx,booktabs,caption}
\usepackage[flushleft]{threeparttable}
\usepackage{bm}
\usepackage[caption=false,font=normalsize,labelfont=sf,textfont=sf]{subfig}
\usepackage{multicol}
\usepackage{multirow}
\usepackage{caption}
\def\muxlsr{{$\tt SAMU\text{-}XLS\text{-}R$}}
\def\mbart{{$\tt MBART$}}
\def\xlsr{{$\tt XLS\text{-}R$}}

\def\labse{{$\tt LaBSE$}}

\def\y{\mathbf{y}}
\def\YY{\mathcal{Y}}
\def\TT{\mathcal{T}}
\def\f{\mathbf{f}}

\def\XX{\mathcal{X}}

\def\a{\mathbf{a}}

\def\e{\mathbf{e}}

\def\z{\mathbf{z}}
\def\ctxt{\mathbf{c}}

\begin{document}

\title{Improved Cross-Lingual Transfer Learning For Automatic Speech Translation}

\author{Sameer Khurana\thanks{Sameer Khurana (email: skhurana@mit.edu), Nauman Dawalatabad, and James Glass are with MIT Computer Science and Artificial Intelligence Laboratory, USA. Luis Vicente, Pablo Gimeno, and Victoria Mingote are with the University of Zaragoza, Spain. Antoine Laurent is with LIUM University, France.
This work was partially performed using HPC resources from GENCI–IDRIS, grants AD011012527 and AD011012565. 
This work has received funding from the European Union’s Horizon 2020 research and innovation programme under the Marie Skłodowska-Curie grant agreement No 101007666.},
        Nauman Dawalatabad,
        Antoine Laurent,
        Luis Vicente,
        Pablo Gimeno,
        Victoria Mingote,\\
        James Glass
}

\markboth{Preprint}%
{Shell \MakeLowercase{\textit{et al.}}: A Sample Article Using IEEEtran.cls for IEEE Journals}


\maketitle

\begin{abstract}
Research in multilingual speech-to-text translation is topical. Having a single  model that supports multiple translation tasks is desirable. The goal of this work it to improve cross-lingual transfer learning in multilingual speech-to-text translation via semantic knowledge distillation. We show that by initializing the encoder of the encoder-decoder sequence-to-sequence translation model with \muxlsr{}, a multilingual speech transformer encoder trained using multi-modal (speech-text) semantic knowledge distillation, we achieve significantly better cross-lingual task knowledge transfer than the baseline \xlsr{}, a multilingual speech transformer encoder trained via self-supervised learning. We demonstrate the effectiveness of our approach on two popular datasets, namely, CoVoST-2 and Europarl.  On the 21 translation tasks of the CoVoST-2 benchmark, we achieve an average improvement of 12.8 BLEU points over the baselines. In the zero-shot translation scenario, we achieve an average gain of 18.8 and 11.9 average BLEU points on unseen medium and low-resource languages. We make similar observations on Europarl speech translation benchmark. 

\end{abstract}


\section{Introduction}
Self-Supervised Representation Learning (SSRL) from speech \cite{harwath2016unsupervised, hsu2017unsupervised, 8683131, pascual2019learning, schneider2019, Baevski2020vqwav2vec:, chung2020generative, khurana2020convolutional, Baevski2020wav2vec,Harwath2020Learning, conneau2020unsupervised, Liu_2021, hsu2021hubert, w2v_bert, babu2021xlsr, alex_cross_modal, samu_xls_r, mslam} has improved tremendously over the past few years due to the introduction of Contrastive Predictive Coding (CPC) \cite{cpc}, a self-supervised representation learning method applied to speech, text, and visual data. The introduction of the core idea of \textit{noise contrastive estimation} \cite{pmlr-v9-gutmann10a} in CPC has led to a series of papers in speech SSRL, such as Wav2Vec \cite{schneider2019}, VQ-Wav2Vec \cite{Baevski2020vqwav2vec:}, Wav2Vec-2.0 \cite{Baevski2020wav2vec}, Multilingual Wav2Vec-2.0 \cite{conneau2020unsupervised}, XLS-R (\xlsr{}, a bigger version of the multilingual wav2vec-2.0) \cite{babu2021xlsr}. Pre-trained SSRL speech encoders like \xlsr{} are considered "foundation models" \cite{foundation_models} for downstream multilingual speech processing applications such as Multilingual Automatic Speech Recognition \cite{conneau2020unsupervised, rivire2020unsupervised, babu2021xlsr}, Multilingual Speech Translation \cite{fst_eff_ft, babu2021xlsr, mslam}, and other para-linguistic property prediction tasks \cite{shor_para, yang2021superb}. This work focuses on Multilingual Speech Translation.

Multilingual Speech Translation (MST) refers to translating speech in all the source languages in set $\XX{}$ to text in all the target languages in set $\YY{}$, which implies a total of $|\TT{}| = |\XX{}| \times |\YY{}|$ translation tasks. In MST, we train a single model for all the translation tasks given by set $\TT{}$. The benefits of having a single model instead of individual model for each task $t \in \TT{}$ are two-fold: First, it is convenient to maintain and share a single model that can perform multiple tasks rather than having $|\TT{}|$ separate models, and second, sharing model parameters amongst $|\TT{}|$ translation tasks could lead to knowledge transfer across tasks, especially from high-resource to low-resource. 

The standard neural network architecture used for MST is the \textit{encoder-decoder} model \cite{ilya_seq_to_seq,vaswani2017attention}. Recently, MST has seen significant improvements owing to; (i) better initialization of the translation model's encoder and decoder with pre-trained speech encoders, like \xlsr{} \cite{babu2021xlsr}, and text decoders, like \mbart{} \cite{liu-etal-2020-multilingual-denoising}, (ii) better fine-tuning strategies \cite{fst_eff_ft}, and (iii) parallel speech-text translation corpora \cite{europarl_st, wang2020covost}. However, as we demonstrate in Section~\ref{sec:cl_gap}, the performance on low-resource tasks remains poor, and in particular, the performance gap (cross-lingual transfer gap) between high and low-resource languages remains large. We hypothesize that this is because the \xlsr{} speech encoder learns non-robust \textit{surface-level features} from unlabeled speech data, rather than the high-level linguistic knowledge about \textit{semantics}. 

To inject semantic knowledge into the learned \xlsr{} representations, we turn to the recently introduced 
Semantically-Aligned Multimodal Cross-Lingual Representation Learning framework, \muxlsr{} \cite{samu_xls_r}. \muxlsr{} (Section~\ref{sec:samu}) is a \textit{knowledge-distillation} framework that distills semantic knowledge from a pre-trained text embedding model into the pre-trained multilingual \xlsr{} speech encoder. The KD framework outputs the \muxlsr{} speech encoder that learns a semantically structured multilingual speech manifold, where a spoken utterance is close to its spoken translations in several other languages that are present in the \muxlsr{}'s training pool (Details in Section~\ref{sec:samu}).  
We \textbf{claim} that building MST model using the semantic representations learned by \muxlsr{} would lead to better cross-lingual transfer from high to low-resource tasks compared to the \xlsr{} and other non-semantic multilingual speech encoder baselines. We verify our claim through several experiments (Section~\ref{sec:results}).
Through this work, we make the following \textbf{contributions}:
\begin{itemize}
    \item We doubled the number of languages previously supported by \muxlsr{} encoder from 25 to 53 (Section~\ref{sec:samu_extend}).
    \item \muxlsr{} encoder provides speech embedding at the utterance level. Previous works \cite{https://doi.org/10.48550/arxiv.2210.05291} have explored the use \muxlsr{} embedding for automatic spoken language understanding and semantic speech-text and speech-speech retrieval \cite{samu_xls_r}. Differently, in this work, for the first time, we show that the fine-grained contextual embeddings (corresponding to 20ms duration speech segment) that precede the coarse utterance level representation (3-10s long speech segment) learned by \muxlsr{} is well-suited for the sequence generation task of multilingual speech translation. We empirically demonstrate, through several experiments, the superiority of \muxlsr{} representations over \xlsr{} and other multilingual speech encoders (Section~\ref{sec:results}).
    \item On the public CoVoST-2 $\XX{}\rightarrow\text{English}$ MST benchmark \cite{wang2020covost}, we show that by switching the \xlsr{} encoder in the MST model to the \muxlsr{} encoder, the performance improves significantly on medium-resource $\XX{}\rightarrow\text{English}$ language group by 15.6 BLEU points, and on low-resource language group by 18.9 BLEU points, and overall by 13.8 BLEU points (Section~\ref{sec:covost_results}).
    \item We also show the efficacy of the \muxlsr{} speech encoder in Zero-Shot translation settings. 
    In the Zero-Shot MST scenario, where we train the \muxlsr{}-based MST model only on high-resource $\XX{}$$\rightarrow$English tasks in CoVoST-2, we observe an improvement on medium and low-resource tasks of 18.8 and 11.9 BLEU points over the \xlsr{} baseline (Section~\ref{sec:zero_x_en}).
    \item Finally, we extend our studies on another MST benchmark, namely Europarl \cite{europarl_st}, which consists of several $\XX{}$$\rightarrow$$\YY{}$ translation tasks. In the zero-shot MST scenario, we observe an overall improvement of 8.5 BLEU points average with \muxlsr{} encoder over \xlsr{}, owing to the significant increase in performance on unseen (during training) source languages of 17 BLEU points (Section~\ref{sec:euro_results}).
\end{itemize}

\section{Motivation: Cross-Lingual Transfer Gap}
\label{sec:cl_gap}
To motivate our work, we show the performance of the multilingual \xlsr{} speech encoder on the CoVoST-2 speech-to-text translation benchmark \cite{wang2020covost}. CoVoST-2 comprises 21 X$\rightarrow$EN speech-to-text translation tasks, where X refers to the language of speech utterance, and EN refers to the corresponding English text translation. \xlsr{} speech encoder is pre-trained via Self-Supervised Learning using unlabeled speech data in 128 languages. Babu et al. \cite{babu2021xlsr} fine-tunes the pre-trained \xlsr{} encoder combined with a pre-trained \mbart{} text decoder \cite{liu-etal-2020-multilingual-denoising} simultaneously on 21 X$\rightarrow$EN translation tasks (multi-task learning) in CoVoST-2 benchmark. We categorize the 21 translation tasks into high, mid, and low-resource groups. A task is classified as high if it has more than 100 hours of paired speech(X)-text(EN) translation training data, mid if training data is between 10 and 100 hours, and low if training data is less than 10 hours. There are four high, five mid, and 12 low-resource tasks.
\begin{figure}[h]
    \centering
    \caption{We report translation performance on 21 X$\rightarrow$EN speech-to-text translation tasks in CoVoST-2 benchmark with different sized pre-trained XLS-R encoders fine-tuned on labeled speech translation data. The 21 tasks are categorized into high, mid, and low resource tasks depending on the available labeled training data for a task. We report average BLEU-4 scores in the three categories. The important thing to consider is the performance gap or cross-lingual transfer gap between high and low-resource translation tasks. We address this large gap in this paper.}
    \includegraphics[width=\linewidth]{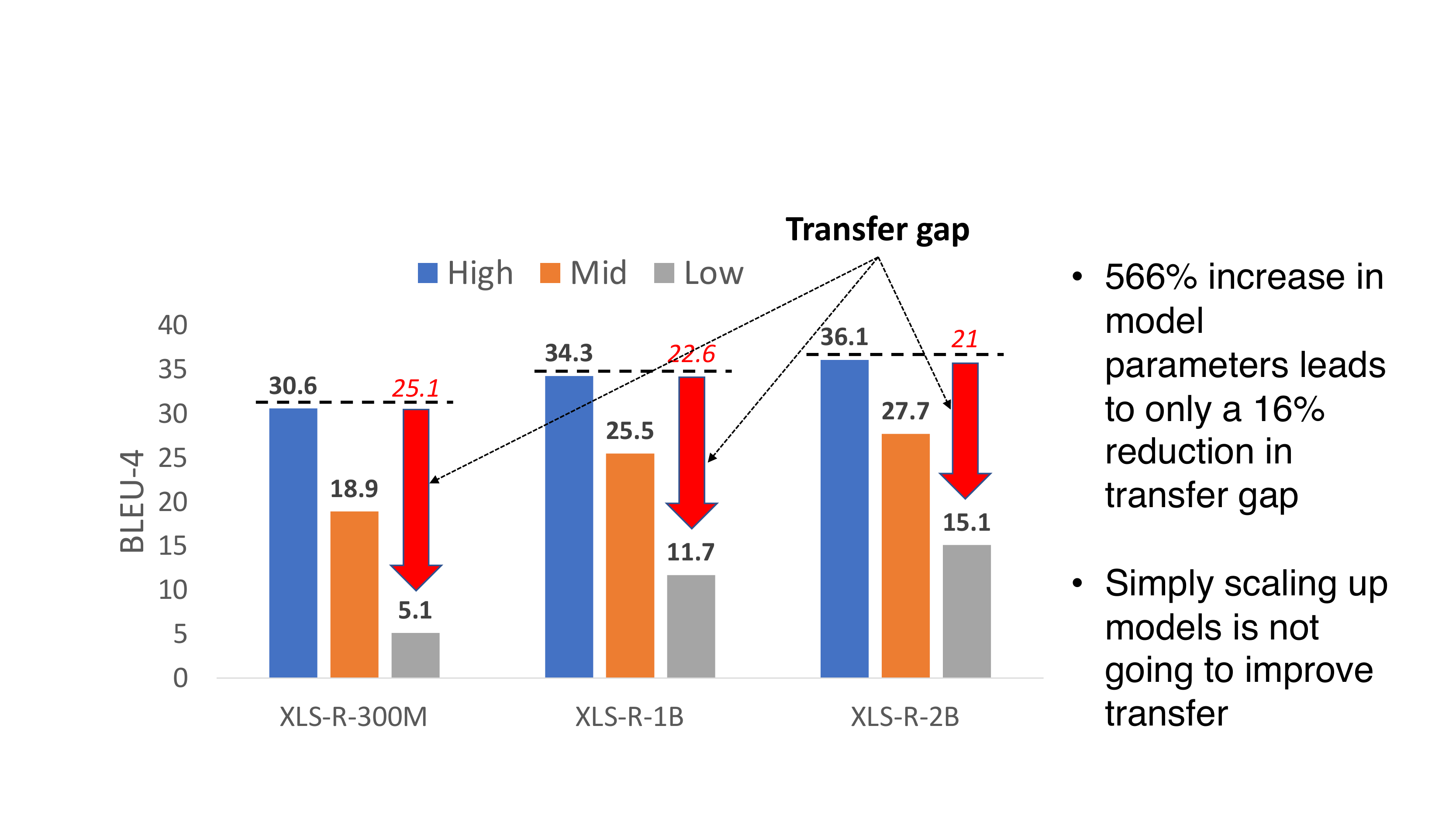}
    \label{fig:motivation_translation}
\end{figure}
\begin{figure*}
    \centering
    \caption{\muxlsr{} \textit{semantic knowledge-distillation} framework. The learning framework comprises a speech and a text encoder. The speech encoder transforms a raw speech waveform into an embedding vector. The text encoder transforms the transcript corresponding to the speech utterance into an embedding. The text encoder is initialized using the pre-trained Language-Agnostic BERT Sentence Embedding model \labse{} \cite{feng2020languageagnostic}. The speech encoder below the pooling layer is initialized using the pre-trained \xlsr{} speech encoder \cite{babu2021xlsr}.}
    \includegraphics[width=0.8\linewidth]{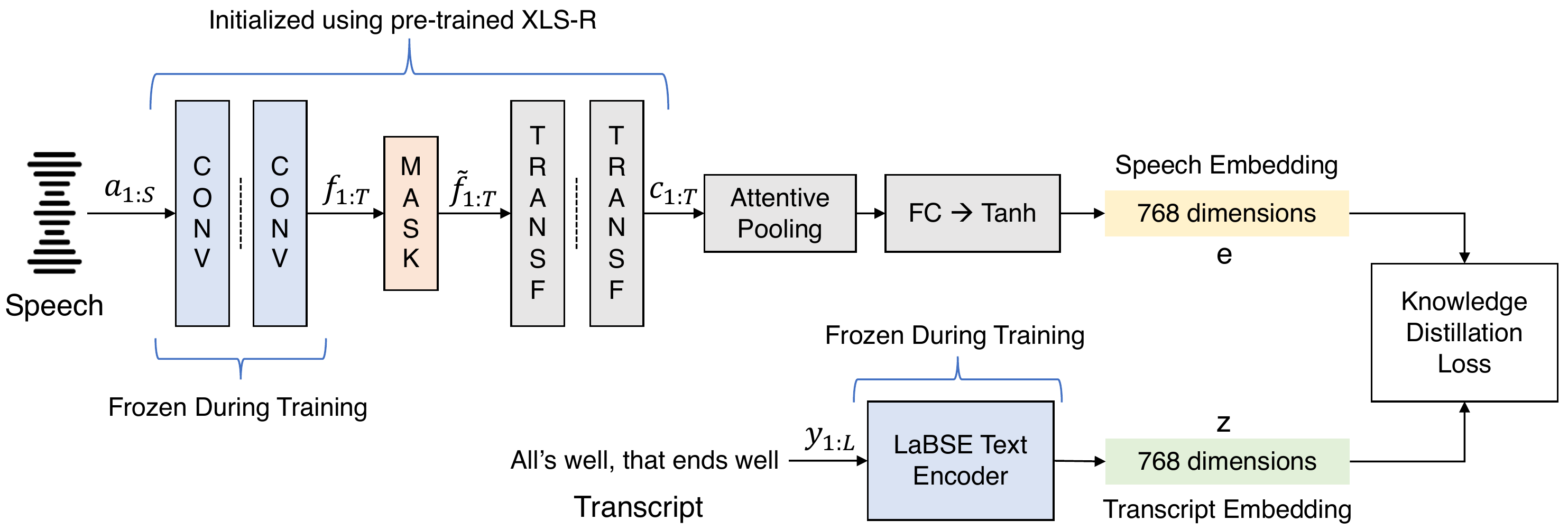}
    \label{fig:samu_training}
\end{figure*}

We report the \xlsr{} (speech encoder)$\rightarrow$\mbart{} (text decoder) transformer model's performance on the high, mid, and low-resource translation groups in Fig.~\ref{fig:motivation_translation}. We report the average BLEU-4 score on each translation group. The vital thing to observe is the performance gap (cross-lingual transfer gap) between high and low-resource translation groups for different-sized \xlsr{} encoders ranging from 300M to 2B parameters. So, increasing the model size from 300M to 2B, a more than 500\% increase leads to only a 16\% reduction in the cross-lingual transfer gap. Since the translation model is built on top of the pre-trained \xlsr{} encoder's representations, there is some missing ingredient in the pre-trained representations that leads to a poor cross-lingual transfer from high to low-resource tasks. This implies that the knowledge that the \xlsr{}-based translation model acquires while learning to perform high-resource X$\rightarrow$EN translation tasks is not useful (or transferable) for learning the low-resource tasks. 

We \textbf{claim} that the missing piece is semantic knowledge. We hypothesize that since \muxlsr{} is specifically trained to encode semantic information in its internal representations, building a translation model on top of \muxlsr{} would lead to better cross-lingual transfer from high to low-resource translation tasks, thus reducing the cross-lingual transfer gap mentioned above.
\section{Preliminaries}
\subsection{XLS-R (\xlsr{})}
This section discusses the architecture of the \xlsr{} encoder. The \xlsr{} encoder consists of a Convolutional Neural Network (CNN) \cite{726791} feature extractor, denoted by $h$, and a transformer encoder \cite{vaswani2017attention}, denoted by $g$.

\subsubsection{CNN Feature Extractor} $h:\a_{1:S}\rightarrow\f{}_{1:T}$ maps the speech waveform $(\a_{1:S} \mid \a_{s} \in \mathbb{R})$ to an intermediate representation $(\f_{1:T}\mid \f_{t} \in \mathbb{R}^{d})$, where $T=S/r$, where $r$ is the factor by which $h$ downsamples $\a_{1:S}$, and $d$ is the size of the feature dimension. For the \xlsr{} encoder, $r=320$, and $d=1024$. The feature extractor $h$ consists of seven temporal Convolution layers ({\tt CONV}). Each layer uses ${\tt GeLU}$ activation function \cite{gelu}. The feature maps outputted by each layer have $512$ channels. The output of the last $\tt CONV$ layer in $h$ is up-projected using a fully-connected layer ($\tt FC$) so that the feature vector $f_t$ has the same dimension as the transformer encoder. Note that the receptive field of each feature vector is 20ms of the input speech waveform. The feature extractor is followed by another $\tt CONV$ layer that encodes the relative position of the feature sequence $\f_{1:T}$, which is used as input to the transformer encoder $g$.

\subsubsection{Transformer Encoder} $g:\f_{1:T}\rightarrow\ctxt{}_{1:T}$ maps $(\f_{1:T} \mid \f_{t} \in \mathbb{R}^{d})$ to contextualized representation $(\ctxt{}_{1:T} \mid \ctxt{}_{t} \in \mathbb{R}^{d})$. The receptive field of each context vector $\ctxt{}_{t}$ is equal to the length of the input speech waveform $\a_{1:S}$, which is usually 5-10s. The transformer encoder consists of 24 layers. Each transformer layer consists of Multi-Headed Self-Attention ($\tt MHSA$), followed by two $\tt FC$ layers. Layer Normalization ($\tt LN$) is used at the input of $\tt MHSA$ and $\tt FC$ blocks. Residual connections adds the input of $\tt MHSA$, with its output, and the input of $\tt FC$ block with its output. The transformer encoder layer dimension is 1024 (size of the feature vector outputted by a transformer layer), and the $\tt FC$ layer dimension is 3072. Each $\tt MHSA$ block consists of 16 attention heads. The output of the first $\tt FC$ layer is processed by a $\tt ReLU$ non-linearity \cite{relu}.

\subsubsection{Training Details} \xlsr{} is pre-trained on approximately 400K hours of multilingual \textit{unlabeled} speech data segmented into utterances of size 3-10s. The training data consists of unlabeled speech in 128 languages coming from the following corpora: VoxPopuli \cite{wang-etal-2021-voxpopuli}, CommonVoice (CoVo) \cite{commonvoice}, Multilingual Speech (MLS) \cite{Pratap2020MLSAL}, BABEL, and Voxlingua \cite{voxling107}. For full training details, see \cite{babu2021xlsr}. We use the pre-trained \xlsr{} checkpoints publicly released\footnote{\url{https://github.com/facebookresearch/fairseq/tree/main/examples/wav2vec/xlsr}}.

As baselines for multilingual X$\rightarrow$EN speech translation experiments (Section~\ref{sec:covost_results}), we use the \xlsr{}$\rightarrow$\mbart{} translation models that are trained on CoVoST-2 X$\rightarrow$EN MST benchmark, officially released here\footnote{\url{https://huggingface.co/facebook/wav2vec2-xls-r-300m-21-to-en}}. The official release comprises three translation models corresponding to 0.3B, 1B, and 2B parameter \xlsr{} speech encoders. The text decoder in all three cases in the 400M parameter \mbart{} decoder.

\begin{table*}[h!]
    \centering
    \caption{The number of hours of transcribed speech data available for training \muxlsr{} in each of the 53 languages from the CommonVoice-Version8 corpus.}
    \setstretch{1.2}
    \begin{tabular}{ccccccccccc}\toprule
         ar&be&bg&ca&cs&cy&da&de&el&en&eo\\\midrule
         85.2&903.9&8.2&916.8&54.9&116.3&6.6&1062.8&15.9&2185.8&1407.9\\\hline
         es&et&eu&fa&fi&fr&fy-NL&ga-IE&gl&ha&hi\\\midrule
         404.6&33.0&98.9&317.3&8.5&826.1&49.6&4.3&10.2&3.4&11.7\\\hline
         hu&id&it&ja&ka&kmr&ky&lt&lv&mn&mt\\\midrule
         19.9&25.8&310.6&40.8&7.6&47.0&37.2&17.4&7.1&12.4&8.3\\\hline
         nl&pl&pt&ro&ru&rw&sk&sl&sv-SE&sw&ta\\\midrule
         98.0&142.2&112.0&15.8&162.6&2000.7&17.7&9.6&40.8&146.8&217.7\\\hline
         th&tr&tt&ug&uk&uz&vi&zh-CN&zh-HK&zh-TW\\\midrule
         142.1&65.1&29.2&59.8&63.4&81.0&4.5&68.0&99.7&62.6
    \end{tabular}
    \label{tab:samu_extend_data}
\end{table*}
\subsection{SAMU-XLS-R (\muxlsr{})}
\label{sec:samu}
\xlsr{} encoder's representation manifold encodes low-level linguistic knowledge, as evident by the non-existent cross-lingual semantic speech retrieval capability when using \xlsr{}'s learned representations in \cite{samu_xls_r}. \muxlsr{} is a bi-modal (speech \& text) \textit{knowledge distillation} (KD) framework for expanding the representation space of \xlsr{} speech encoder also to encode high-level semantic knowledge. The \muxlsr{} learning framework uses multilingual transcribed speech set $\mathcal{D} = \{\a^n_{1:S}, \y^n_{1:L}\}_{i=1}^{N}$ consisting of speech, $\a^i_{1:S}$ paired with its text transcript $\y^i_{1:L}$. Note that speech and its transcript are in the same language. The \muxlsr{} KD framework consists of a speech and a text processing branch as illustrated in Figure~\ref{fig:samu_training}, and detailed below.

\subsubsection{The Speech Branch} The speech branch maps the speech waveform $\a_{1:S}$ of duration 3-10s to a single embedding vector $\e \in \mathbb{R}^{768}$. The mapping is performed in the following two steps: (i) The pre-trained \xlsr{} speech encoder maps $\a_{1:S}$ to the contextual representation $\ctxt{}_{1:T}$, and (ii) An attention-based \textit{temporal pooling} \cite{safari2020self}, and a non-linear ($\tt tanh$) down-projection layer transforms $\ctxt{}_{1:T}$ to a single embedding $\e$ of size 768.

\subsubsection{The Text Branch} The text branch consists of a pre-trained \labse{} sentence embedding encoder \cite{feng2020languageagnostic} that transforms the text transcript $\y_{1:L}$ to a semantic embedding $\z \in \mathbb{R}^{d}$, of size $d=768$. 
\labse{} supports 109 written languages. It embeds sentences from different languages in a shared semantic embedding space, i.e., a sentence and its translation lie close together in the \labse{} embedding space. 
By regressing on the \labse{} semantic embedding, the \xlsr{} speech encoder learns to encode semantics hidden in the spoken utterance in its internal representations. This new \textit{semantics-aware} speech encoder is referred to as \muxlsr{} encoder.

\subsubsection{Training Details.} The \muxlsr{} encoder is trained to minimize the cosine distance between the speech and the text embedding $\e$, and $\z$ respectively (see Figure~\ref{fig:samu_training}). The \textit{loss function} is given by:
\begin{equation}
    \mathcal{L} = \beta * \left(1.0 - \frac{\e \cdot \z}{||\e||\ ||\z||}\right)
\end{equation}
Where $\beta$ is used to scale up the magnitude of the cosine distance loss, a small loss magnitude could lead to extremely small gradient update values and significantly longer training times. Upscaling the loss by a constant $\beta$ could alleviate this problem. The parameters of the speech branch are tuned to minimize the above-mentioned loss $\mathcal{L}$. The text encoder \labse{} remains fixed during training along with the feature extractor $h$ of the pre-trained \xlsr{} encoder (See Figure~\ref{fig:samu_training} and \cite{samu_xls_r} for training details).

\section{Method}
\subsection{Expanding SAMU-XLS-R}
\label{sec:samu_extend}
We double the number of languages supported by \muxlsr{} to 53 than the 25 supported previously. For training \muxlsr{}, we use speech utterances from multiple languages annotated with their text transcripts in the same language. We refer to \muxlsr{}'s training data pool as Background (BKG) data. BKG data depends on the downstream MST task domain. Ideally, we would like BKG data from the same domain as the MST task (in-domain). We use the multilingual transcribed speech from the CommonVoice-Version8 (CoVo-V8) corpus \cite{commonvoice}. CoVo-V8 consists of transcribed speech in 87 languages (26
language families). Around 53 languages overlap with the language set supported by Language-Agnostic BERT Sentence Encoder (LaBSE), which provides semantic
supervision for training the SAMU-XLS-R speech encoder. The 53 languages are: Arabic (ar), Belarusian (be), Bulgarian (bg), Catalan (ca), Czech (cs), Welsh (cy), Danish (da), German (de), Greek (el), English (en), Estonian (eo), Spanish (es), Estonian (et), Basque (eu), Persian (fa), Finnish (fi), French (fr), Western Frisian (fy-NL), Gaelic (ga-IE), Galician (gl), Hausa (ha), Hindi (hi), Hungarian (hu), Indonesian (id), Italian (it), Japanese (ja), Georgian (ka), Khmer (kmr), Kyrgyz (ky), Lithuanian (lt), Latvian (lv), Mongolian (mn), Maltese (mt), Dutch (nl), Polish (pl), Portuguese (pt), Romanian (ro), Russian (ru), Kinyarwanda (rw), Slovak (sk), Slovenian (sl), Swedish (sv-SE), Swahili (sw), Tamil (ta), Thai (th), Turkish (tr), Tatar (tt), Uyghur (ug), Ukaranian (uk), Uzbek (uz), Vietnamese (vi), Chinese-Mandarin (zh-CN), Chinese-HongKong (zh-HK), Chinese-Taiwanese (zh-TW).
Table~\ref{tab:samu_extend_data} presents the number of hours of transcribed speech available in each of the 53 languages for \muxlsr{} training. The total training hours is 12.7K. BKG data is highly imbalanced, where the majority of the data comes from a few high-resource languages. Following \cite[Equation 3]{samu_xls_r}, we balance the BKG data to have a proportionate representation of different languages in a training mini-batch, which avoids under-fitting to low-resource languages. The up-sampling process involves repeating utterances in low-resource languages. Unlike previous work, we use offline data augmentation, namely speed perturbation \cite{ko15_interspeech}, to improve the robustness of the learned representations to the data distribution shifts.

\subsection{Translation Model}
\subsubsection{Overview} We use the standard encoder-decoder architecture for our translation model.
We initialize the encoder using pre-trained \muxlsr{}. Following \cite{fst_eff_ft},  the decoder is initialized with the decoder of a pre-trained text-to-text translation model, namely \mbart{}\footnote{ \url{https://github.com/facebookresearch/fairseq/tree/main/examples/multilingual\#mbart50-models}}. The encoder-decoder model is trained using corpora that consist of tuples $(\a_{1:S}, \y_{1:L})$, where $\y_{1:L}$ is the text translation sequence of the speech sequence $\a_{1:S}$.

\subsubsection{Training} Given a speech-text translation pair $(\a_{1:S}, \y_{1:L})$, we tune the parameters of the translation model to maximize the log-probability
of $\y_{1:L}$ conditioned on speech utterance $\a_{1:S}$ (log-likelihood). Also, we employ \textit{label-smoothing} \cite{label_smooth, ls_hinton}, where a ground-truth $\y_l$ label is randomly replaced with the label predicted by the model $\hat{\y}_l$. We set the probability of token replacement as 0.3.
We use the Adam optimizer \cite{adam_opt} to maximize the log-likelihood with the optimal (according to a development set) learning rate (lr) of 5e-4. We use a three-phase learning rate scheduler \cite{Baevski2020wav2vec} as follows; (i) warm-up lr to 5e-4 for the first 10\% of the training iterations, (ii) keep the lr constant for the next 40\% of iterations, and (iii) decay the lr linearly for the rest of training. We use 28K training iterations and train the model on 8 A100 GPUs. A single training iteration uses a batch size of 10 minutes of speech utterances paired with their text translations. We use \textit{mixed-precision} training style \cite{mixed_precision}; most computations are performed on half-precision floating point numbers except the final loss computation. We use the \textit{fairseq} toolkit \cite{fairseq} for model training.

Following \cite{Baevski2020wav2vec}, we mask (time and feature dimension) the feature sequence $\f_{1:T}$, output by the CNN feature extractor of \muxlsr{}. 
The masking process is performed in the following two steps: (i) with some probability, referred to as the \textit{masking probability}, choose a masking index, and (ii) mask $M$ consecutive indices starting from the chosen index. $M$ is known as the \textit{masking span}. For the time dimension, we set 0.3 as the masking probability and the mask span of six. For the feature dimension, we set the masking probability to 0.5 and 64 as the mask span. The time and feature masking parameters are chosen according to a development set. The above-mentioned data augmentation is akin to SpecAugment \cite{park2019specaugment}, a commonly used speech spectrogram data augmentation method for training speech-to-text generation models.

The translation model comprises 700 million trainable parameters (300M encoder and 400M decoder parameters). We fine-tune only 75 million parameters. Most encoder parameters are fixed to the pre-trained \muxlsr{} parameters, and the decoder parameters are fixed to the pre-trained \mbart{} decoder. Below we give details about encoder and decoder fine-tuning.

\subsubsection{Encoder Fine-Tuning} \label{subsec:encoder_ft} We keep all the parameters of the speech encoder fixed to their pre-trained \muxlsr{} values. Instead, we add a small number of downstream \textit{task-specific} parameters to each transformer layer of the speech encoder following \cite{Houlsby2019ParameterEfficientTL}. We insert adapters, a \textit{bottleneck} Feed-Forward layer, after the Multi-Headed Self-Attention ($\tt MHSA$) and Fully-Connected ($\tt FC$) blocks in each transformer layer. 
An adapter comprises a single hidden layer with $\tt ReLU$ activation. The input and the output layers have the same size. In contrast, the hidden layer is a fraction of the input layer size. We found the optimal (according to a development set) size of the hidden layer to be a fourth of the input layer size. 

The motivation for using adapters is two-fold: (i) \textit{Parameter Efficiency:} We only tune the adapter layer parameters, a fraction of the total parameters that comprise the speech encoder. In our case, the \muxlsr{} encoder consists of 300M parameters, and we add 75M task-specific adapter parameters. Hence, we only fine-tune the 75M parameters on the downstream task. (ii) \textit{Avoids Knowledge Forgetting:} By freezing the encoder's parameters to pre-trained values, we preserve the essential semantic knowledge encoded in the \muxlsr{} encoder that it acquired during the semantic knowledge-distillation pre-training phase. We later show (Section~\ref{subsec:adapt_vs_full}) that preserving this semantic knowledge is essential to achieving good cross-lingual transfer. Hence, adapters form a crucial part of our cross-lingual transfer learning strategy. It must be noted that research on adapters is ever-expanding. An in-depth analysis of specific adapter choices on downstream task performance is out-of-scope for this paper.

\subsubsection{Decoder Fine-Tuning} Like the encoder, we keep most parameters fixed to their pre-trained values acquired during the multilingual text-to-text translation task. Unlike the encoder, we do not introduce any new task-specific parameters. Instead, we tune a fraction of the decoder's parameters. Each layer in the transformer decoder consists of the Masked (causal) Multi-Headed Self-Attention ($\tt MMHSA$), Encoder-Decoder Cross-Attention ($\tt CA$), and two Fully-Connected ($\tt FC$) blocks. The inputs to the $\tt MMHSA$, $\tt CA$, and $\tt FC$ blocks are normalized via Layer Normalization ($\tt LN$) method presented in \cite{ba2016layer}.
Following \cite{fst_eff_ft}, we only tune the parameters of the $\tt LN$ and $\tt CA$ blocks. We fine-tune $\tt CA$ because, previously, it is trained as part of a decoder in a text-to-text translation pipeline. Hence, the $\tt CA$ module needs to be fine-tuned again for the downstream MST task to make it amenable to the input from the speech encoder. Moreover, we fine-tune $\tt LN$ because it is task and dataset-specific and empirically improves the performance of MST \cite{fst_eff_ft}.

\subsubsection{Inference}
We use \textit{beam search}, with a beam size of 5, to generate text translations for a given speech utterance. The inference process is \textit{offline}, i.e., the decoder takes the full input speech utterance into account to generate the output translation. The decoder generates translation in an autoregressive manner. We do not use any \textit{external language model} during inference.

\section{Evaluation}
\subsection{Translation Scenarios}
\label{sec:trans_scenario}
We tackle the following translation scenarios in this work.
\subsubsection{Multilingual Translation}
We simultaneously train a translation model on several speech-to-text translation tasks in this scenario. E.g., we train a single model for all 21 X$\rightarrow$EN translation tasks in the CoVoST-2 benchmark. Most of the training data come from a few high-resource translation tasks such as FR$\rightarrow$EN and DE$\rightarrow$EN, while most tasks such as ID$\rightarrow$EN are low-resource. In this scenario, we compare different translation models to test for cross-lingual translation task transfer from high to low-resource translation tasks.

\subsubsection{Zero-Shot Multilingual Translation}
Given a set of translation tasks, we train a translation model on a subset of the tasks while keeping the rest hidden during training in this translation scenario. E.g., we train an X$\rightarrow$EN speech-to-text translation model using high-resource translation tasks in the CoVoST-2 benchmark while keeping the mid and low-resource tasks unseen during training. We compare translation models for zero-shot cross-lingual task transfer in this scenario from high to mid and low-resource X$\rightarrow$EN translation tasks.

\subsection{Translation Tasks}
\label{sec:trans_tasks}
\begin{figure}[t]
    \centering
    \caption{Number of hours of labeled training data (Y-Axis) for all the 21 X$\rightarrow$EN translation tasks in the CoVoST-2 benchmark.}
    \includegraphics[width=\linewidth]{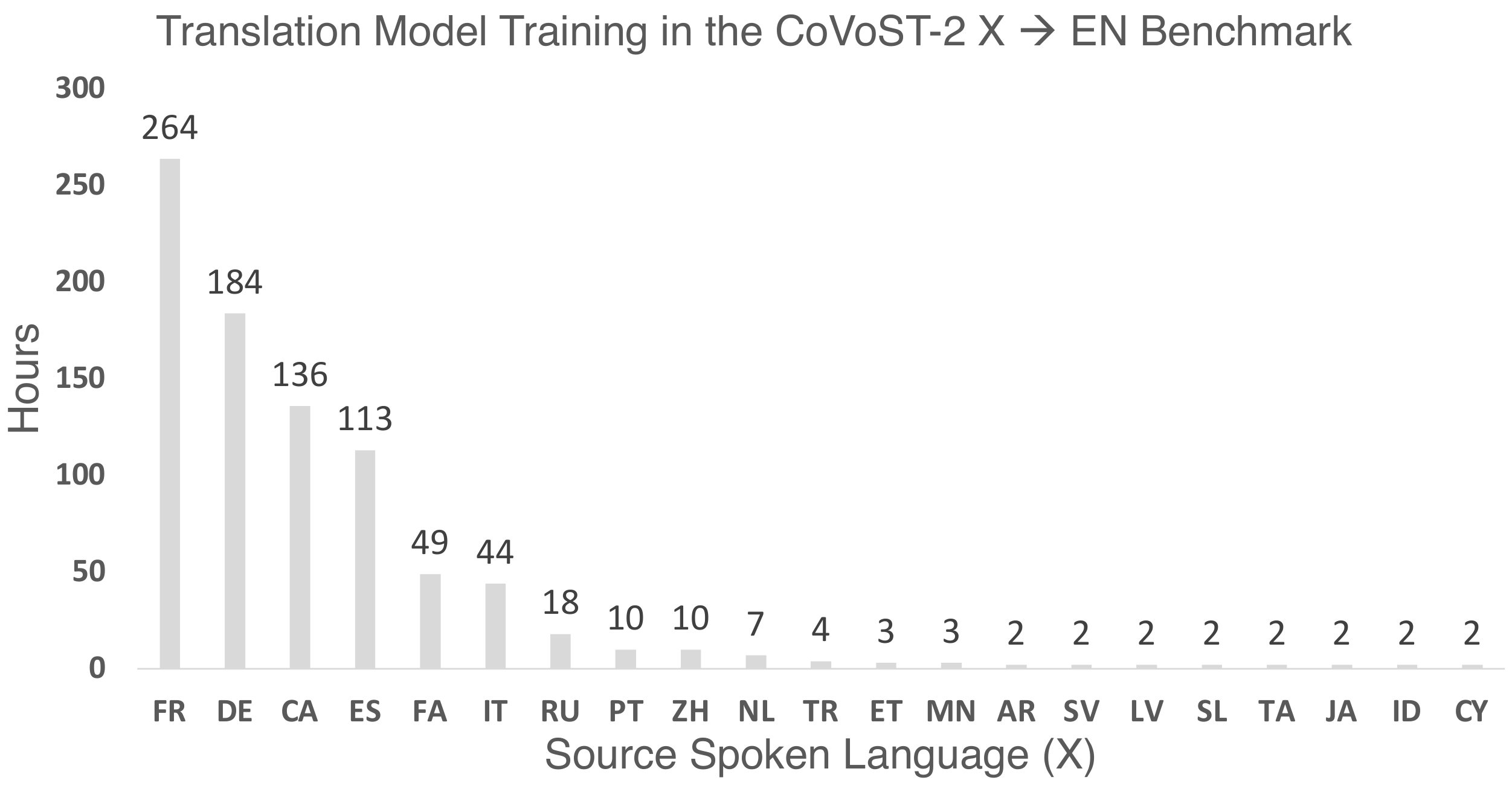}
    \label{fig:covost_data_x_en}
\end{figure}
\begin{table}[t]
    \centering
    \caption{Training data (hours) for the 72 translation tasks X$\rightarrow$Y in the Europarl Speech-to-Text translation benchmark.}
    \renewcommand{\arraystretch}{1.2}
    \resizebox{0.99\linewidth}{!}{%
    \begin{tabular}{c|ccccccccc}
         SRC/TGT&FR&DE&IT&ES&PT&PL&RO&NL&EN  \\\midrule
         FR&-& 21&20&21&22&20&18&22&32\\
         DE&18&-&17&18&18&17&17&18&30\\
         IT&21&21&-&21&21&21&19&20&37\\
         ES&14&14&14&-&14&13&12&13&22\\
         PT&10&10&10&10&-&9&9&9&15\\
         PL&18&18&17&18&18&-&16&18&28\\
         RO&12&12&12&12&12&12&-&12&24\\
         NL&5&5&4&5&4&4&4&-&7\\
         EN&81&83&80&81&81&79&72&80&-
    \end{tabular}
    }
    \label{tab:europarl_data}
\end{table}
We build translation models to tackle the following translation tasks in this work.
\subsubsection{X$\rightarrow$EN Speech-to-Text Translation} We build translation models for the 21 X(Speech)$\rightarrow$EN(Text) translation tasks in the CoVoST-2 translation benchmark \cite{wang2020covost}. The 21 spoken languages in CoVoST-2 are French (fr), German (de), Spanish (es), Catalan (ca), Italian (it), Russian (ru), Chinese (zh), Portuguese (pt), Persian (fa), Estonian (et), Mongolian (mn), Dutch (nl), Turkish (tr), Arabic (ar), Swedish (sv), Latvian (lv), Slovenian (sl), Tamil (ta), Japanese (ja), Indonesian (id), and Welsh (cy). The 21 translation tasks are divided into high, mid, and low-resource groups, depending on the amount of labeled data available for a translation task. High-resource tasks have more than 100 hours of labeled training data, mid-resource tasks have between 10 and 100 hours, and low-resource have less than ten hours of labeled training data. CoVoST-2 has four high-resource translation tasks corresponding to fr, de, es, and ca source languages and five mid-resource tasks corresponding to it, ru, zh, pt, and fa languages. The rest of the tasks are low-resource. Fig.~\ref{fig:covost_data_x_en} presents the training data for each of the 21 translation tasks. Notice the data imbalance among different tasks. 


In the multilingual translation scenario, we simultaneously train translation models on all 21 tasks mentioned above. We only train translation models on the four high-resource tasks in the zero-shot scenario.

\begin{table*}[h!]
    \centering
    \caption{We compare our proposed \muxlsr{}-300M translation model with several other translation models, whose encoders are initialized using differently sized pre-trained \xlsr{} multilingual unimodal speech encoders. The performance is measured using BLEU-4, Google-BLEU, ROUGE-L, METEOR, BERTScore, and NIST translation metrics.}
    \begin{tabular}{lrrrrcrrrr}\toprule
         &\multicolumn{4}{c}{BLEU-4}&&\multicolumn{4}{c}{Google-BLEU}\\
         \cmidrule{2-5}\cmidrule{7-10}
         Model&High&Mid&Low&TRFGap&&High&Mid&Low&TRFGap\\\midrule
         \xlsr{}-300M&30.6&18.9&5.1&25.1&&0.36&0.24&0.10&0.26\\
         \xlsr{}-1B&34.3&25.5&11.7&22.6&&0.38&0.29&0.16&0.22\\
         \xlsr{}-2B&36.1&27.7&15.1&21.0&&0.39&0.31&0.20&0.19\\
         \muxlsr{}-300M&34.4&31.1&20.3&14.1&&0.38&0.34&0.24&0.13\\
         {\tt Cascaded}&32.6&29.7&22.5&10.1&&0.36&0.33&0.26&0.10\\
         {\tt Transcripts}&36.4&34.2&27.9&8.5&&0.39&0.37&0.32&0.08\\\midrule
         &\multicolumn{4}{c}{ROUGE-L}&&\multicolumn{4}{c}{METEOR}\\
         \cmidrule{2-5}\cmidrule{7-10}
         Model&High&Mid&Low&TRFGap&&High&Mid&Low&TRFGap\\\midrule
         \xlsr{}-300M&0.60&0.44&0.23&0.37&&0.62&0.45&0.24&0.38\\
         \xlsr{}-1B&0.61&0.49&0.31&0.30&&0.63&0.50&0.32&0.31\\
         \xlsr{}-2B&0.63&0.53&0.38&0.25&&0.65&0.53&0.39&0.26\\
         \muxlsr{}-300M&0.62&0.58&0.42&0.19&&0.64&0.59&0.45&0.19\\
         {\tt Cascaded}&0.60&0.56&0.45&0.14&&0.61&0.56&0.46&0.15\\
         {\tt Transcripts}&0.64&0.61&0.52&0.11&&0.66&0.62&0.54&0.12\\\midrule
         &\multicolumn{4}{c}{BERTScore}&&\multicolumn{4}{c}{NIST}\\
         \cmidrule{2-5}\cmidrule{7-10}
         Model&High&Mid&Low&TRFGap&&High&Mid&Low&TRFGap\\\midrule
         \xlsr{}-300M&0.56&0.33&0.04&0.52&&8.0&5.0&1.9&6.1\\
         \xlsr{}-1B&0.58&0.41&0.16&0.42&&8.3&5.9&3.0&5.3\\
         \xlsr{}-2B&0.61&0.45&0.25&0.36&&8.6&6.3&3.7&4.9\\
         \muxlsr{}-300M&0.59&0.54&0.34&0.25&&8.3&7.1&4.4&4.0\\
         {\tt Cascaded}&0.56&0.50&0.37&0.20&&8.2&6.9&4.9&3.3\\
         {\tt Transcripts}&0.62&0.58&0.47&0.14&&8.7&7.7&5.7&3.0
    \end{tabular}
    \label{tab:results_x_en_covost2}
\end{table*}
\begin{table}[h!]
    \centering
    \caption{We compare our proposed \muxlsr{}-300M translation model with ${\tt mSLAM}$ translation models, whose encoders are initialized using differently sized pre-trained ${\tt mSLAM}$ multilingual multimodal speech encoders. The performance is measured using the BLEU-4 translation metric.}
    \begin{tabular}{lrrrrcrrrr}\toprule
         &\multicolumn{4}{c}{BLEU-4}&\\
         \cmidrule{2-5}
         Model&High&Mid&Low&TRFGap\\\midrule
         ${\tt mSLAM}$-600M&37.6&27.8&15.1&22.5\\
         ${\tt mSLAM}$-2B&37.8&29.6&18.5&19.3\\
         \muxlsr{}-300M&34.4&31.1&20.3&14.1
    \end{tabular}
    \label{tab:results_x_en_covost2_mslam}
\end{table}
\begin{table*}[h!]
    \centering
    \caption{We compare our proposed \muxlsr{}-300M translation model with several other translation models, whose encoders are initialized using differently sized pre-trained \xlsr{} multilingual unimodal speech encoders. The performance is measured using the BLEU-4 translation metric.}
    \begin{tabular}{lrrrrrrrrr}\toprule
         Model&fr&de&es&ca&it&fa&ru&zh&pt\\\midrule
         \xlsr{}-300M&34.9&29.3&35.9&30.6&30.9&6.3&30.0&5.2&30.8\\
         \xlsr{}-1B&36.3&31.4&37.8&32.1&33.5&9.1&35.7&6.9&41.4\\
         \xlsr{}-2B&37.6&33.6&39.1&33.9&35.0&13.0&39.5&9.4&41.8\\
         \muxlsr{}-300M&36.1&31.7&37.9&31.9&34.0&22.0&42.1&13.1&44.2\\
         {\tt Cascaded}&32.7&30.9&36.4&30.3&32.5&14.3&42.9&14.4&44.2\\
         {\tt Transcripts}&38.4&34.1&41.1&32.1&36.4&24.2&46.0&17.4&46.9\\\midrule
         Model&nl&tr&et&mn&ar&sv&lv&sl&ta\\\midrule
         \xlsr{}-300M&25.1&6.7&4.0&0.2&3.9&11.3&7.4&8.3&0.0\\
         \xlsr{}-1B&29.6&11.1&8.0&0.6&9.3&24.5&15.7&16.8&0.1\\
         \xlsr{}-2B&31.6&16.9&11.2&1.5&17.1&29.7&19.7&19.0&0.5\\
         \muxlsr{}-300M&34.9&28.4&11.6&3.4&36.6&28.5&1.9&12.9&4.0\\
         {\tt Cascaded}&33.0&25.1&17.4&0.0&35.7&39.6&20.3&27.8&1.6\\
         {\tt Transcripts}&36.3&28.8&25.6&3.8&44.6&46.4&29.2&38.4&2.2\\\midrule
         Model&cy&ja&id\\\midrule
         \xlsr{}-300M&2.8&0.6&1.2\\
         \xlsr{}-1B&6.61&1.3&7.8\\
         \xlsr{}-2B&14.2&3.5&16.4\\
         \muxlsr{}-300M&34.1&13.1&34.4\\
         {\tt Cascaded}&6.4&19.4&43.6\\
         {\tt Transcripts}&9.0&20.8&49.8\\
         
    \end{tabular}
    \label{tab:results_x_en_covost2_langwise}
\end{table*}
\subsubsection{X$\rightarrow$Y Speech-to-Text Translation} We develop translation models for the 72 X$\rightarrow$Y translation tasks in the Europarl benchmark \cite{europarl_st}. There are nine spoken languages in Europarl namely, en, fr, de, it, es, pt, pl, ro, and nl. Speech utterances in each spoken language are paired with their corresponding text translations in eight other languages. In the zero-shot scenario, we train the translation models on 32 tasks corresponding to the following four source languages fr, de, es, and it. Each of the four source languages is paired with eight target languages. Table~\ref{tab:europarl_data} shows each translation task's labeled training data. 

\subsection{Translation Models}
\label{sec:base_top_models}
\subsubsection{\muxlsr{}-300M} We propose \muxlsr{}-300M transformer model for translation in this work. The encoder is initialized using the pre-trained \muxlsr{} speech encoder (Section~\ref{sec:samu_extend}), and the decoder is initialized using the pre-trained \mbart{} text decoder. The suffix 300M in \muxlsr{}-300M refers to the model's size of 300M parameters.

\subsubsection{\xlsr{}-(300M, 1B, 2B)} We compare \muxlsr{}-300M with three \xlsr{} speech encoder based translation models namely, \xlsr{}-300M, \xlsr{}-1B, and \xlsr{}-2B. The three translation models differ from \muxlsr{}-300M model in that the encoder of the translation model is initialized using pre-trained \xlsr{} speech encoders of different sizes ranging from 300M to 2B parameters. The decoder is initialized using the pre-trained \mbart{} decoder. Unlike our multimodal \muxlsr{} speech encoder, \xlsr{} is only trained using unlabeled speech data. Also, \muxlsr{} is specifically trained to learn semantic representations, while \xlsr{} has no constraints imposed during its training phase to encode semantic knowledge.

\subsubsection{mSLAM} We compare \muxlsr{}-300M translation model against two $\tt mSLAM$ \cite{mslam} speech encoder based translation models namely, $\tt mSLAM$-600M, and $\tt mSLAM$-2B. Like \muxlsr{}, $\tt mSLAM$ speech encoder is a multimodal speech-text encoder. Unlike \muxlsr{}, which is trained using semantic supervision from a pre-trained semantic text encoder, $\tt mSLAM$ is not trained with explicit semantic supervision.

\subsubsection{Cascaded Translation} We compare \muxlsr{}-300M with a strong \textit{cascaded} translation system. We perform the translation in two steps: (i) Transcribe the speech utterance using an ASR model, and (ii) Use a text-to-text translation model to translate the ASR transcript to text in a target language. We use whisper-large-v2 \footnote{\url{https://huggingface.co/openai/whisper-large-v2}} \cite{radfordrobust} as the ASR model in the first step and \mbart{} \cite{liu-etal-2020-multilingual-denoising} text-to-text translation model for the second step in the cascade. For X$\rightarrow$EN cascade, we use \mbart{}-many-to-English text-to-text translation model\footnote{\url{https://huggingface.co/facebook/mbart-large-50-many-to-one-mmt}}. The Whisper ASR model is multilingual that supports transcription of around 93 languages. Since \cite{radfordrobust} shows that whisper achieves state-of-the-art ASR performance on several public benchmarks, we choose Whisper for automatically transcribing speech. \mbart{}-many-to-English translation model can translate text from 50 languages to English.

\subsubsection{Transcripts} As a topline, we use the ground-truth text transcripts corresponding to speech utterances and use \mbart{}-many-to-English to translate to English.

\section{Results}
\label{sec:results}
\subsection{Multilingual X$\rightarrow$EN Translation} \label{sec:covost_results}
\textbf{Table~\ref{tab:results_x_en_covost2}} shows the performance of different translation models on the high, mid, and low-resource translation groups in the CoVoST-2 speech-to-text translation benchmark. We compare our proposed \muxlsr{}-300M translation model against \xlsr{}-300M, \xlsr{}-1B, and \xlsr{}-2B translation models. CoVoST-2 comprises 21 X$\rightarrow$EN translation tasks, and the translation models are trained simultaneously on all translation tasks. See Section~\ref{sec:trans_tasks} for details about the translation tasks, and Section~\ref{sec:base_top_models} for details about the different translation models.

The model performance is measured using the standard translation metrics, namely BLEU-4 \cite{post-2018-call}, Google-BLEU \cite{wu2016googles}, ROUGE-L \cite{lin-2004-rouge}, METEOR \cite{banarjee2005}, BERTScore \cite{eddine2021frugalscore}, and NIST \cite{10.5555/1289189.1289273}. We make the following \textbf{observations}: (i) On \textbf{High resource tasks}, the \xlsr{}-2B model performs the best, with SAMU-XLS-R-300M lagging a couple of points behind. Compared to the similar-sized \xlsr{}-300M model, \muxlsr{}-300M performs 4 BLEU points better. (ii) On \textbf{Mid resource tasks}, \muxlsr{}-300M outperforms all the models achieving a BLEU score of 31.1, which is significantly better than \xlsr{}-300M model's BLEU score of 5.1. \muxlsr{}-300M even outperforms the much larger \xlsr{}-2B speech encoder by 3.3 BLEU points. (iii) On \textbf{Low resource tasks}, \muxlsr{}-300M performs the best. Compared to the similar-sized \xlsr{}-300M model, \muxlsr{}-300M does better by 15 BLEU points. It also outperforms the much larger \xlsr{}-2B by 5.2 BLEU points. The cross-lingual transfer gap (TRFGap), which is the difference in performance between high and low resource task groups, is significantly less (14.1 BLEU) for \muxlsr{}-300M model compared to other models. Second to \muxlsr{}-300M is \xlsr{}-2B, which has a TRFGap of 21 BLEU points while having 500\% more parameters. Similar arguments can be made using metrics other than BLEU-4. On all metrics, \muxlsr{} is significantly better than the \xlsr{} baselines and comparable to the strong cascaded system, while significantly worse than the model which uses the human transcripts corresponding to speech utterances for translation into English using the \mbart{}-many-to-English text translation model. For the subsequent experiments, we only report BLEU-4 scores.

\textbf{Table~\ref{tab:results_x_en_covost2_mslam}} compares \muxlsr{}-300M translation model with ${\tt mSLAM}$-600M, and ${\tt mSLAM}$-2B models that use differently sized pre-trained multimodal (speech-text) multilingual ${\tt mSLAM}$ speech encoder. \muxlsr{}-300M performs better on mid- and low-resource translation tasks. Importantly, \muxlsr{}-300M has a lower cross-lingual transfer gap (TRFGap) between high and low resource groups of 14.1 BLEU points compared to 22.5 for ${\tt mSLAM}$-600M, and 19.3 for ${\tt mSLAM}$-2B. The BLEU scores for ${\tt mSLAM}$ models are lifted from \cite{mslam}. Since mSLAM \cite{mslam} report only BLEU scores on the CoVoST-2 benchmark, and we do not have access to ${\tt mSLAM}$ models, we can not evaluate the model using metrics other than BLEU-4.

The above observations validate our claims that building translation technology on top of semantic speech representations would increase the cross-lingual task knowledge transfer from high to low-resource languages. Similar inferences can be reached by using metrics other than BLEU-4.

\textbf{Table~\ref{tab:results_x_en_covost2_langwise}} shows the performance of different translation models on each of the 21 X$\rightarrow$EN speech-to-text translation tasks in the CoVoST-2 benchmark. We observe that \muxlsr{}-300M significantly outperforms the similarly sized \xlsr{}-300M translation model on several mid and low-resource languages. Some notable improvements are for the following source languages: id (34.4 vs. 1.2 BLEU), cy (34.1 vs. 2.8 BLEU), ja (13.1 vs 0.6 BLEU), sv (28.5 vs 11.3 BLEU), tr (28.4 vs 6.7 BLEU), ar (36.6 vs. 3.9 BLEU), fa (22.0 vs. 6.3 BLEU), and pt (44.2 vs. 30.8 BLEU). For some source languages \muxlsr{}-300M does not do well, such as lv (1.9 BLEU), sl (12.9 BLEU), mn (3.4 BLEU), et (11.6 BLEU), ta (4.0), ja (13.1 BLEU), and zh (13.1 BLEU). For ta, ja, mn, and zh, the topline (${\tt Transcripts}$) model also performs poorly. For sl, and lv, the topline text-to-text translation and cascaded models perform significantly better than the \muxlsr{}-300M model. 

Poor performance of \muxlsr{}-300M on lv, mn, and sl can be explained by the lack of available transcribed speech data in these languages for multimodal pre-training of \muxlsr{} (Section~\ref{sec:samu_extend}). We have 7 hours for lv, 12 hours for mn, and 9 hours of transcribed speech for sl, compared to 317 hours for fa, 85 hours for ar, 116 hours for cy, 98 hours for nl, 40 hours for sv, 25.8 hours for id, 162 hours for ru, 400 hours for es, and close to 1K hours for fr, de, and ca. However, for ta, zh, and ja we have a decent amount of transcribed speech, but still, the performance is relatively poor. This can be explained away by observing the performance of the topline (${\tt Transcripts}$), where we use a pre-trained \mbart{}-many-to-English text-to-text translation model (Details in Section~\ref{sec:base_top_models}) that translates the ground-truth transcripts corresponding to speech utterances in language X to text in English. The topline performance for zh, ja, and ta is relatively poor. Since we use the decoder of \mbart{}-many-to-English model to initialize the decoder of our speech-to-text translation model \muxlsr{}-300M, we also observe poor speech-text-translation performance on these tasks.

\subsection{Zero-Shot X$\rightarrow$EN Speech-to-Text Translation.}
\label{sec:zero_x_en}
\begin{figure}[h!]
    \centering
    \caption{We report average BLEU-4 for the zero-shot X$\rightarrow$EN multilingual speech-to-text translation scenario on the high, mid, and low resource task groups in the CoVoST-2 benchmark. We compare our translation model \muxlsr{}-300M with the similarly sized \xlsr{}-300M translation model. The translation models are only trained on high-resource groups, while the mid and low-resource groups are \textit{unseen} during training.}
    \includegraphics[width=\linewidth]{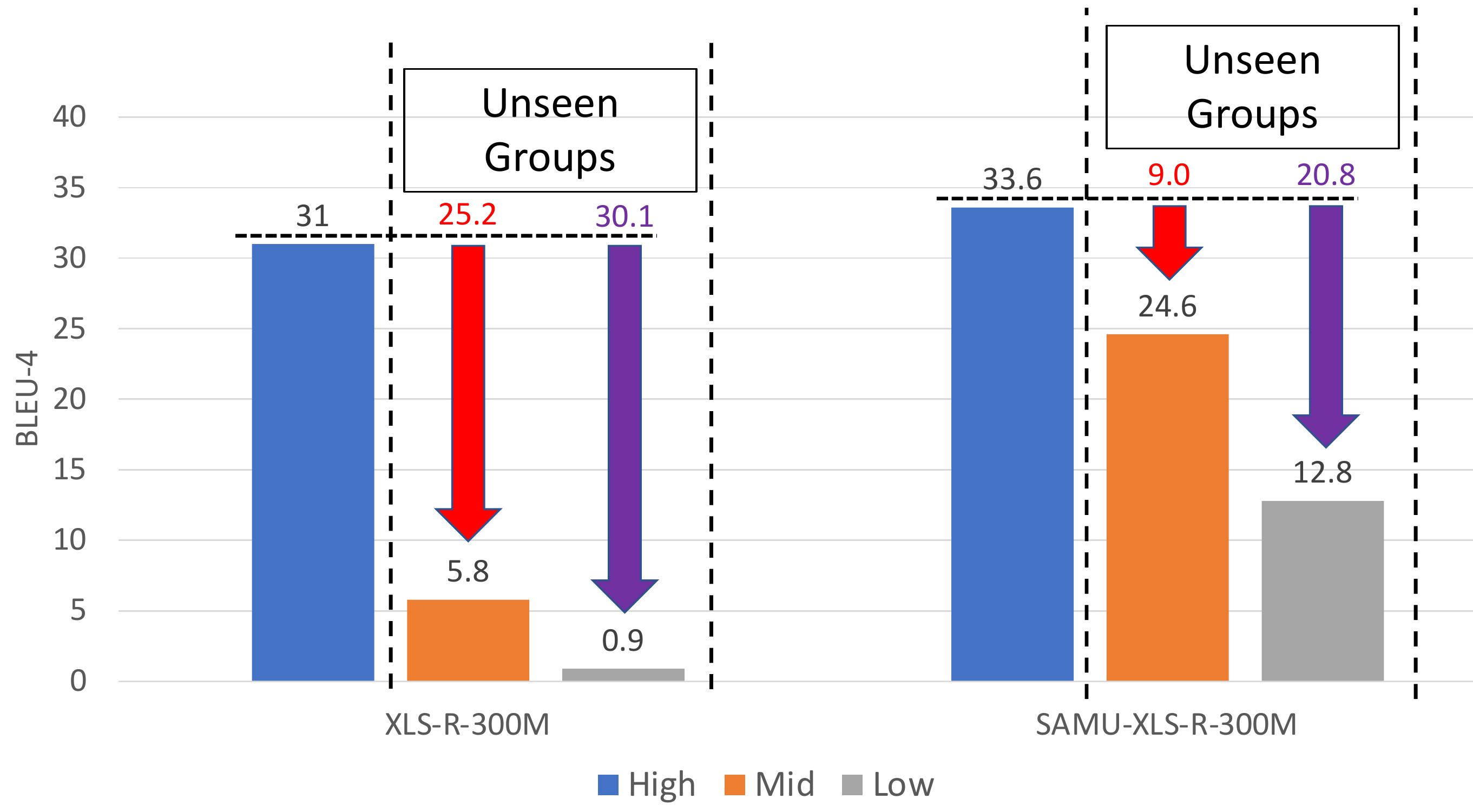}
    \label{fig:covost_x_en_zero}
\end{figure}

Next, we train the translation models on four high-resource X$\rightarrow$EN translation tasks in the CoVoST-2 benchmark (See Section~\ref{sec:trans_tasks} for details). We evaluate the X$\rightarrow$EN translation models on the high, mid, and low task groups to test for zero-shot cross-lingual transfer capability of \muxlsr{}-300M from high to mid and low-resource X$\rightarrow$EN tasks. We compare \muxlsr{}-300 with \xlsr{}-300M translation model. Figure~\ref{fig:covost_x_en_zero} presents the results. We observe that \muxlsr{}-300M performs on average 18.8 BLEU points better in the mid-resource and 11.9 BLEU points in the low-resource group. The cross-lingual transfer gap between the high \& mid and high \& low groups is significantly smaller for \muxlsr{}-300M (9.0, and 20.8) than \xlsr{}-300M (25.2, and 30.1).

\begin{figure*}[h!]
    \centering
    \caption{Absolute BLEU score improvements using \muxlsr{}-300M over \xlsr{}-300M baseline on the 72 X$\rightarrow$Y translation tasks in the Europarl benchmark. The translation models are trained on a subset of 32 translation tasks, corresponding to four source languages, while 40 tasks are unseen during training corresponding to five source languages.}
    \includegraphics[width=0.8\linewidth]{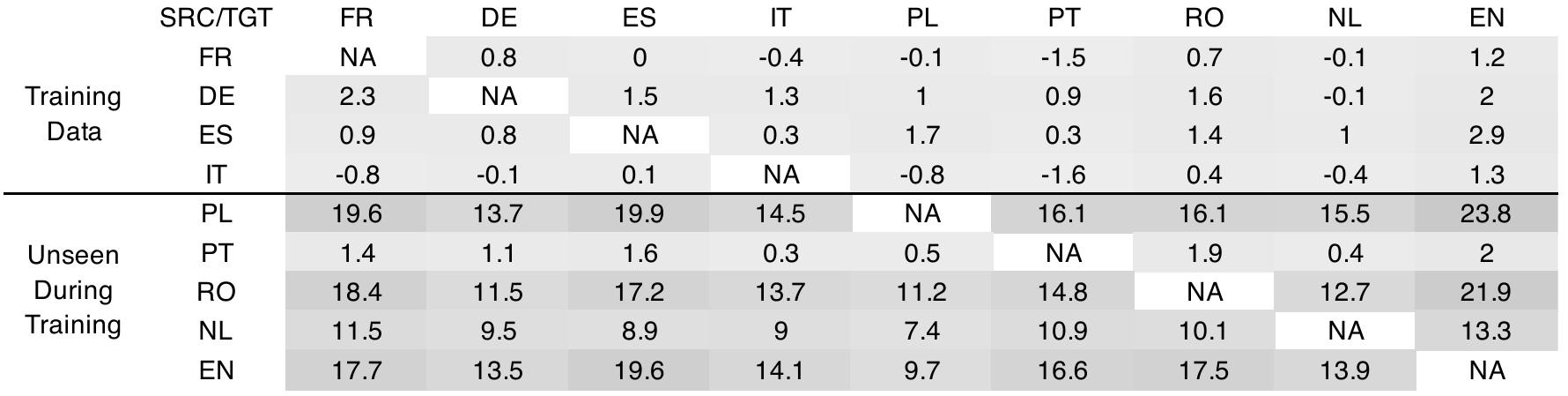}
    \label{fig:samu_europarl_x_to_y_zero}
\end{figure*}
\begin{figure}[h!]
    \centering
    \caption{We compare on the Europarl X$\rightarrow$EN benchmark \xlsr{} and \xlsr{}-CTC initialization of the translation model's encoder. \xlsr{} is pre-trained using unlabeled speech via self-supervised learning, while \xlsr{}-CTC refers to the \xlsr{} encoder that is fine-tuned (after self-supervised pre-training) using transcribed speech data. We report the BLEU-4 score for eight source spoken languages. Speech utterances in each source language are paired with its English text translations.} 
    \includegraphics[width=\linewidth]{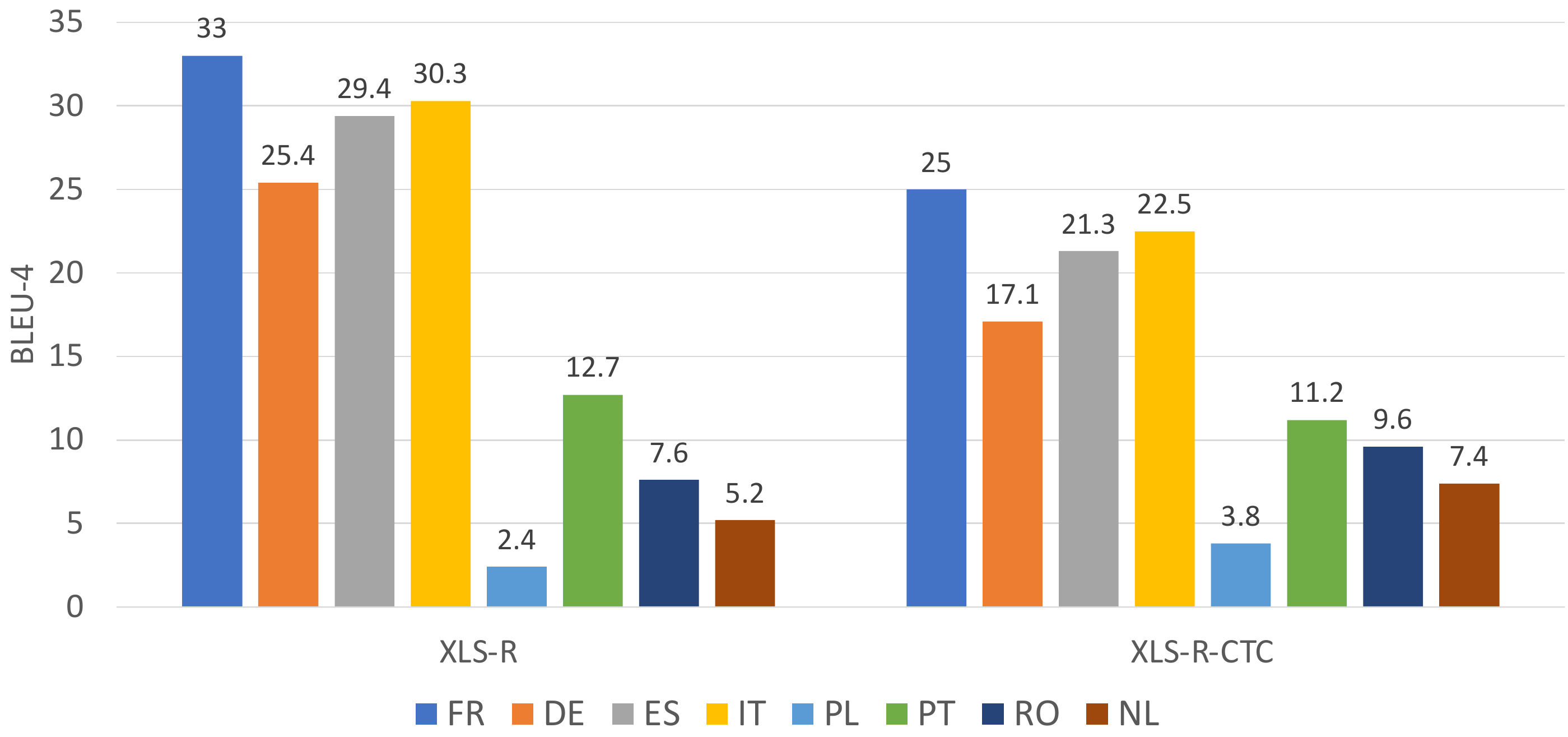}
    \label{fig:xlsr_europarl_ablation_ctc}
\end{figure}
These results strengthen our claims that building speech translation technology with semantic speech representations would improve cross-lingual transfer across languages. Note that zero-shot implies that the translation model during its training does not see any paired X$\rightarrow$EN translation data for mid and low-resource languages X. Transcribed speech data was available for these languages during multimodal pre-training of \muxlsr{} speech encoder which is used to initialize the encoder of the \muxlsr{}-300M translation~model.

\subsection{Zero-Shot X$\rightarrow$Y Speech-to-Text Translation.} \label{sec:euro_results} Finally, to further bolster our claims about the usefulness of semantic speech representations for translation, we compare translation models on the 72 X$\rightarrow$Y translation tasks in the Europarl Benchmark. See Section~\ref{sec:trans_tasks} for details about translation tasks and available data for training translation models.

We train translation models \muxlsr{}-300M and \xlsr{}-300M on a 32-task subset out of 72 tasks. The 32 tasks correspond to four source spoken languages: fr, de, es, and it, while the remaining 40 tasks correspond to five source languages: pl, pt, ro, nl, and en. Speech utterances in each source language are paired with text translations in eight other languages. Unlike X$\rightarrow$EN translation models discussed above, we initialize the decoder of X$\rightarrow$Y translation models with the decoder of \mbart{}-many-to-many\footnote{\url{https://huggingface.co/facebook/mbart-large-50-many-to-many-mmt}} model instead of \mbart{}-many-to-English, since we have to generate translations in multiple target languages. 

Figure~\ref{fig:samu_europarl_x_to_y_zero} compares the performance of \xlsr{}-300M, and \muxlsr{}-300M on the 72 translation tasks. We report the absolute BLEU-4 score improvement that \muxlsr{}-300M achieves over \xlsr{}-300M baseline translation model. The darker the cell in the figure, the greater the improvement in the BLEU score. We observe that \muxlsr{}-300M performs drastically better than \xlsr{}-300M on the 40 unseen (during translation model training) translation tasks and at par on the 32 seen (during translation model training) translation tasks. We observe the biggest improvements in unseen tasks such as pl$\rightarrow$en (23.8), ro$\rightarrow$en (21.9), pl$\rightarrow$en (19.9), pl$\rightarrow$fr (19.6), ro$\rightarrow$es (17.2), etc. Overall, \muxlsr{}-300M improves over \xlsr{}-300M baseline model by an average of 12 BLEU-4 points.

\begin{table}[h!]
    \centering
    \caption{We compare the translation model's performance when using Adapter-based fine-tuning vs. fine-tuning all the encoder parameters. The performance is measured using the \textbf{BLEU-4} translation metric.}
    \resizebox{0.99\linewidth}{!}{%
    \setstretch{1.2}
    \begin{tabular}{lrrrrcrrr}\toprule
         Model&fr&de&es&it&pl&pt&ro&nl\\\midrule
         \xlsr{}-\textbf{F}&33.0&25.4&29.4&30.3&2.4&12.7&7.6&5.2\\
         \xlsr{}-\textbf{A}&29.8&22.4&25.5&27.2&3.2&15.6&9.1&8.2\\
         \muxlsr{}-\textbf{F}&33.2&26.3&29.7&30.7&3.0&12.2&9.8&7.0\\
         \muxlsr{}-\textbf{A}&32.3&25.5&29.3&28.7&24.3&18.4&29.9&24.1
    \end{tabular}
    }
    \label{tab:abltation_x_en_europarl}
\end{table}
\subsection{Analysis}
\subsubsection{\xlsr{} vs. \xlsr{}-CTC}
The experiments above compare \muxlsr{} with \xlsr{} multilingual speech encoder. \xlsr{} is trained using multilingual unlabeled speech data in 128 languages via self-supervised learning, while \muxlsr{} fine-tunes pre-trained \xlsr{} using multilingual transcribed speech data via semantic knowledge distillation. But, what if instead of fine-tuning pre-trained \xlsr{} with semantic supervision from text transcriptions like \muxlsr{} does, we fine-tune \xlsr{} on the task of Automatic Speech Recognition using the same multilingual transcribed speech data that was used for \muxlsr{} training? We compare the translation performance of \xlsr{} and \xlsr{}-CTC encoders in Fig.~\ref{fig:xlsr_europarl_ablation_ctc}. \xlsr{}-CTC refers to the encoder we get after supervised CTC-based fine-tuning of pre-trained \xlsr{} using multilingual transcribed speech data. CTC \cite{Graves2006} is a standard framework for training ASR models \textit{end-to-end}. We observe that the pre-trained \xlsr{} encoder performs better than \xlsr{}-CTC encoder. \xlsr{}-CTC encoder is slightly better on some translation tasks. Still, the difference is so small that our previous observations on the efficacy of our proposed semantic speech encoder \muxlsr{} are not impacted, even though we compared \muxlsr{} with the \xlsr{} encoder, which is trained using unlabeled multilingual speech.

\subsubsection{Adapter vs. Full Encoder Fine-Tuning} \label{subsec:adapt_vs_full} As mentioned in Section~\ref{subsec:encoder_ft}, we perform Adapter-based fine-tuning of the translation model's encoder, where we insert new task-specific parameters in the form of adapters in each encoder layer. We only fine-tune adapter layers while keeping the rest of the layers frozen to their pre-trained values. Table~\ref{tab:abltation_x_en_europarl} compares the translation model's performance when using Adapter-based fine-tuning vs. fine-tuning all the encoder parameters. We observe that using adapter fine-tuning with \xlsr{}-300M (\xlsr{}-A) translation model brings performance gains on low-resource tasks such as ro$\rightarrow$en, nl$\rightarrow$en, pt$\rightarrow$en, and pl$\rightarrow$en. In contrast, the performance degrades significantly for higher-resource tasks compared to full encoder fine-tuning (\xlsr{}-F). We observe a similar trend with \muxlsr{}-300M translation model. But, the performance gains for low-resource tasks are drastic with adapter-based fine-tuning of the encoder (\muxlsr{}-A). Again, this is due to our proposed \textit{semantic} speech encoder \muxlsr{}, which results in a significant cross-lingual transfer from high to low-resource translation tasks. This result also shows that preserving semantic knowledge during training for downstream translation tasks, learned by the \muxlsr{} encoder as a result of our multimodal learning framework, is essential.

\section{Conclusion}
This paper addresses the central question of cross-lingual transfer learning in Natural Language Processing. We focus on the problem of multilingual spoken language translation, which we model using the standard encoder-decoder model. We analyze the impact of different encoder initializations on the downstream translation task performance. We show that by initializing the encoder with an encoder that we pre-train using the newly introduced \textit{semantic knowledge distillation framework} \muxlsr{}, we achieve significantly better cross-lingual transfer in the downstream speech-to-text translation task than the baselines. The baseline translation models use the state-of-the-art multilingual pre-trained speech encoder \xlsr{} and others for initialization.

To substantiate our claims, we perform multilingual translation on two public benchmarks, CoVoST-2 and Europarl. On the 21 X$\rightarrow$English CoVoST-2 speech translation tasks, we achieve an average improvement of 12.8 BLEU points. In the zero-shot scenario, where we train the translation model only on the four high-resource languages while keeping the rest 17 languages unseen (during training), we achieve an average improvement of 11.8 BLEU points over the baseline \xlsr{} encoder initialization. In particular, we achieve drastic improvements of 18.8 and 11.9 average BLEU points on medium and low-resource languages, respectively. We made similar observations on the Europarl X$\rightarrow$Y speech-to-text translation benchmark.

Our work has limitations. Currently, training \muxlsr{} requires access to multilingual transcribed data, which could be hard for many spoken languages. Also, the dependence on a pre-trained text encoder hinders expanding \muxlsr{} to more languages. Hence, future work should focus on injecting semantic information via \textit{weakly supervised learning} using unaligned speech and text data and without using the \labse{} text encoder.
\vfill
\bibliography{taslp}
\bibliographystyle{IEEEtran}
\end{document}